\pdfoutput=1

\documentclass[11pt]{article}

\usepackage[preprint]{acl}

\usepackage{times}
\usepackage{latexsym}

\usepackage[T1]{fontenc}

\usepackage[utf8]{inputenc}

\usepackage{microtype}

\usepackage{inconsolata}

\usepackage{color}

\newcommand{\siyuan}[1]{\textcolor{black}{#1}}

\usepackage{amsthm}
\usepackage{amsmath}
\usepackage{amssymb}
\usepackage{caption}
\usepackage{subcaption}
\usepackage{enumerate}
\usepackage{multirow}
\usepackage{graphicx} 
\usepackage{url}
\usepackage{booktabs}
\usepackage{algorithm}
\usepackage{algorithmic}
\usepackage{makecell}
\usepackage{tcolorbox}
\usepackage{enumitem}
\usepackage{multirow}

\usepackage{array}
\usepackage{colortbl}
\usepackage{hhline} 
\usepackage{graphicx}

\newcommand{\checkmarkgreen}{\textcolor{green}{\scalebox{1.5}{\textbf{\checkmark}}}}
\newcommand{\xmarkred}{\textcolor{red}{\scalebox{1.6}{\textbf{$\times$}}}}

\definecolor{diffadd}{rgb}{0.9, 1.0, 0.9} 
\definecolor{diffdel}{rgb}{1.0, 0.9, 0.9} 
\definecolor{errorhighlight}{rgb}{1.0, 0.5, 0.5} 

\newcommand{\hladd}[1]{\setlength{\fboxsep}{1pt}\colorbox{diffadd}{#1}}
\newcommand{\hldel}[1]{\setlength{\fboxsep}{1pt}\colorbox{diffdel}{#1}}
\newcommand{\hlerror}[1]{\textbf{\textcolor{red}{#1}}} 

\title{Benchmark Self-Evolving: \\A Multi-Agent Framework for Dynamic LLM Evaluation}

\author{Siyuan Wang\textsuperscript{\rm 1}\footnotemark[1], Zhuohan Long\textsuperscript{\rm 2}\footnotemark[1]\footnotetext[1]{Equal contribution.}, \\
\bf Zhihao Fan\textsuperscript{\rm 3}, Zhongyu Wei\textsuperscript{\rm 1}, Xuanjing Huang\textsuperscript{\rm 1} \\
\textsuperscript{\rm 1}Fudan University,
\textsuperscript{\rm 2}Tongji University,
\textsuperscript{\rm 3}Alibaba Inc. \\
\texttt{wangsy18@fudan.edu.cn; loongnanshine@gmail.com} \\
}

\begin{document}
\maketitle
\begin{abstract}
This paper presents a benchmark self-evolving framework to dynamically evaluate rapidly advancing Large Language Models (LLMs), aiming for a more accurate assessment of their capabilities and limitations. 
We utilize a multi-agent system to manipulate the context or question of original instances, reframing new evolving instances with high confidence that dynamically extend existing benchmarks. Towards a more scalable, robust and fine-grained evaluation, we implement six reframing operations to construct evolving instances testing LLMs against diverse queries, data noise and probing their problem-solving sub-abilities. 
With this framework, we extend benchmark datasets of four tasks.
Experimental results show a general performance decline in most LLMs against their original results. This decline under our scalable and robust evaluations, alongside our fine-grained evaluation, more accurately reflect models' capabilities. Besides, our framework widens performance discrepancies both between different models and within the same model across various tasks, facilitating more informed model selection for specific tasks
\footnote{Code and data are available at \url{https://github.com/NanshineLoong/Self-Evolving-Benchmark}}. 
\end{abstract}

\section{Introduction}
Recent advancements in Large Language Models (LLMs)~\cite{touvron2023llama, vicuna2023, openai2023gpt4, jiang2023mistral} have demonstrated remarkable performance across various tasks, ranging from text generation to complex problem-solving. The evaluation of LLMs thus has emerged as a crucial area of research~\cite{chang2023survey, espejel2023gpt}. It can provide a comprehensive understanding of the capabilities and limitations in these models, and guide the selection of the most applicable LLM for specific applications. Besides, a systematic assessment of LLMs would inspire further potential improvement.

A multitude of benchmark datasets~\cite{hendrycks2020measuring, liang2022holistic, srivastava2023beyond} have been proposed to evaluate LLMs. However, with the rapid development and emerging abilities of ever-evolving LLMs, these static datasets are increasingly inadequate for a thorough assessment. Besides, the extensive use of data for improving LLMs leads to data contamination issues~\cite{zhou2023don, shi2023detecting}, where in-domain training or even public test data may be inadvertently included during LLM training, resulting in skewed evaluations. 
\begin{figure}[!t]
    \centering
    \includegraphics[width=1.0\columnwidth]{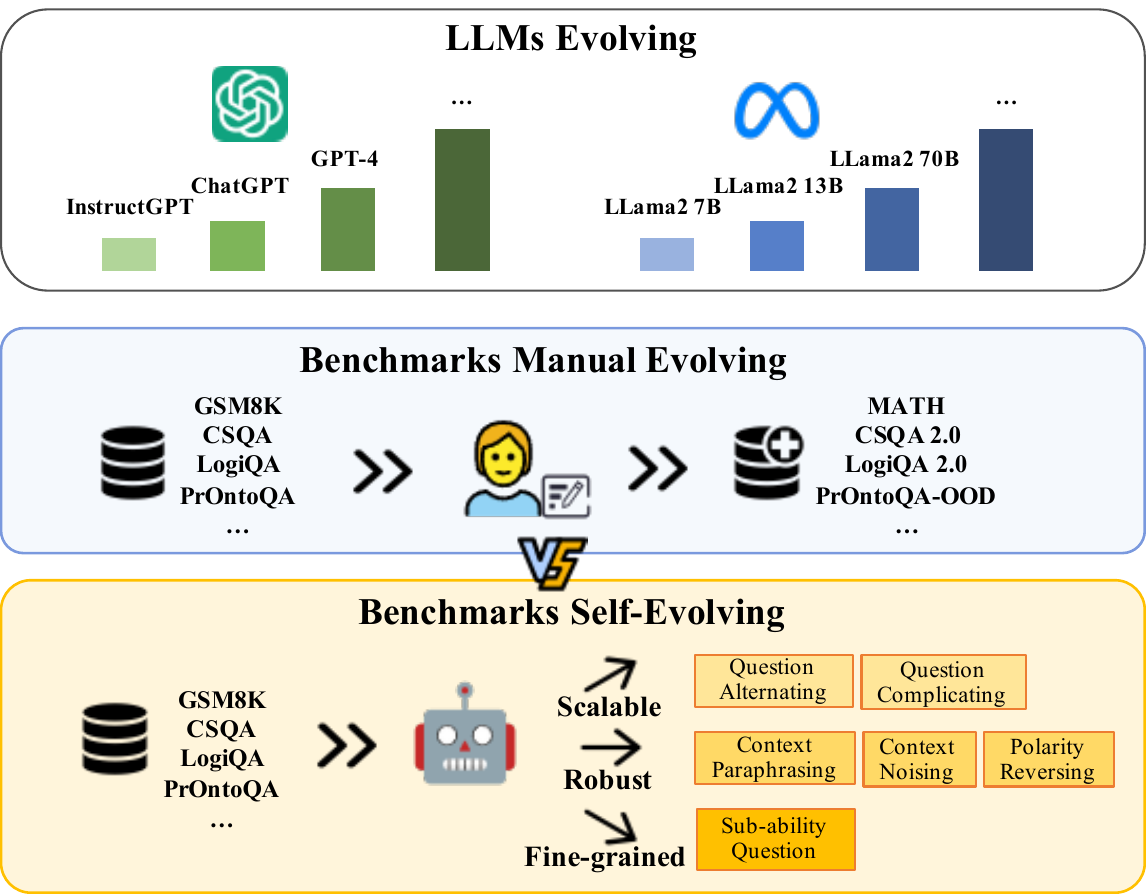}
    \caption{The evolution of LLMs necessitates benchmark self-evolving.}
    \label{fig:system}
    \vspace{-3mm}
\end{figure}
These challenges necessitate continual updates of static benchmark datasets, enabling a more dynamic and accurate evaluation of LLMs. Since annotating new benchmarks from scratch is costly~\cite{kiela2021dynabench}, \citet{wei2023skywork} evaluate LLMs using perplexity on re-sampled data. However, this over-reliance on perplexity may not fully reflect LLMs' performance beyond predictive accuracy.
\citet{zhu2023dyval} dynamically synthetize test samples based on directed acyclic graphs, but this method struggles in generalizing to tasks that cannot be graph-represented. In this work, we propose to flexibly update existing benchmark datasets instead of constructing entirely new ones. 



We introduce a benchmark self-evolving framework, which reframes existing benchmark instances into new variants for dynamic evaluation, by modifying their contexts or questions, and corresponding answers.
This framework propels existing benchmarks towards self-evolution in three directions, providing a systematical dynamic evaluation of LLMs. 
First, to examine LLMs' ability to generalize across diverse and increasingly challenging queries, we introduce \textbf{scalable evaluation} by creating alternative or more complex questions based on original contexts. 
Second, to counteract LLMs' tendency to exploit shortcut biases~\cite{gallegos2023bias, yang2023harnessing} and their sensitivity to data noise~\cite{dong2023revisit, pezeshkpour2023large}, our framework implements \textbf{robust evaluation}. This involves incorporating various perturbations to the contexts of original instances, including paraphrasing, adding noise, and reversing polarity. Finally, to mitigate the impact that outdated data and bias susceptibility could skew capability assessments, we design \textbf{fine-grained evaluation} to probe LLMs’ sub-abilities for solving different problems.

We design a multi-agent system to dynamically generate evolving instances from existing benchmarks while ensuring their accuracy. It comprises four key components: an instance pre-filter, an instance creator, an instance verifier and a candidate option formulator. The workflow begins with the pre-filter to select manageable instances from the original evaluation set. The instance creator crafts new instances by editing their contexts or questions with answers, which the verifier checks for correctness. To further enhance reliability, the candidate option formulator subsequently creates an incorrect answer option for each new context-question pair, which the verifier need to identify as inconsistent with the new context-question. These rigorously generated and double-verified instances will be used for dynamic evaluation. All these components are powered by GPT-4 to leverage its generative and verification strengths.


Based on our framework, we dynamically extend benchmark datasets of four different tasks, including mathematical reasoning (GSM8K), logical reasoning (CLUTRR), commonsense reasoning (StrategyQA) and reading comprehension (BoolQ), and re-evaluate various closed-source and open-source LLMs. Results show that our scalable and robust evaluation are more challenging compared to original benchmarks, leading to a general performance decline for all models. It helps reveal the limited generalizability and robustness of models to diverse and complex queries. This along with sub-ability probing offers a more accurate reflection of LLMs' true capabilities. 
Besides, our framework expands the performance gap between various models and also the differences of a single model across various tasks, which benefits the selection of the most suitable LLM for specific applications.

\section{Benchmark Self-Evolving Framework}
\begin{table*}[!ht]
    \centering
    \resizebox{0.98\textwidth}{!}{
    \begin{tabular}{m{19cm}}
    \midrule
    \midrule
        \centering\arraybackslash \bf Original Instance \\
        \textbf{Context:} Janet's ducks lay 16 eggs per day. She eats three for breakfast every morning and bakes muffins for her friends every day with four. She sells the remainder at the farmers' market daily for \$2 per fresh duck egg. \\
        \textbf{Original Question:} How much in dollars does she make every day at the farmers' market? \\
        \textbf{Original Answer:} Janet sells 16 - 3 - 4 = <<16-3-4=9>>9 duck eggs a day. She makes 9 * 2 = \$<<9*2=18>>18 every day at the farmer's market. \#\#\#\# 18 \\
     \midrule
     \midrule
    \end{tabular}
    }
    \vspace{2mm}

    \resizebox{0.98\textwidth}{!}{
    \begin{tabular}{m{2cm}m{2.1cm}m{1.7cm}m{13.2cm}}
    \toprule
      \centering \bf Directions & \centering \bf Operation Types  & \centering\arraybackslash \bf Changed Items &     \centering\arraybackslash \bf Examples \\
       \midrule
        \centering \multirow{2}{*}[-4ex]{Scalable} & \centering Question Alternating & \centering\arraybackslash context, answer & \textbf{Alternative Question:} If Janet decides to use 2 of her daily eggs to make a special omelette for dinner each day, how much will she earn at the farmers' market in a week? \newline
       \textbf{Alternative Answer:} \$98 \vspace{0.5mm}\\
       \cline{4-4}
       & \centering Question Complicating & \centering\arraybackslash context, answer & \textbf{Complex Question:} How many days will it take for Janet to save \$100 from her earnings at the farmers' market? \newline
       \textbf{Complex Answer:} 6 days \\
       \midrule
       \centering \multirow{3}{*}[-11ex]{Robust} & \centering Context Paraphrasing & \centering\arraybackslash context & \textbf{Paraphrased Context:} Janet's daily egg production from her ducks is 16. Each morning, she consumes three eggs for breakfast and uses four more to bake muffins for her friends. The remaining eggs are then sold at the farmers' market for \$2 each. \vspace{0.5mm}\\
       \cline{4-4}
       & \centering Context Noising & \centering\arraybackslash context & \textbf{Noised Context:} Janet's ducks lay 16 eggs per day and \underline{\emph{her cows product 4L milk}} per day. She eats three eggs and 1L milk for breakfast every morning and bakes muffins for her friends every day with four eggs. She \underline{\emph{keeps the remainder milk for herself}} and only sells the remainder eggs at the farmers' market daily for \$2 per fresh duck egg. \vspace{0.5mm}\\
       \cline{4-4}
       & \centering Polarity Reversing & \centering\arraybackslash context, answer & 
       \textbf{Reversed Context:} Janet's ducks lay \underline{\emph{20 eggs}} per day. She eats \underline{\emph{five}} for breakfast every morning and bakes muffins for her friends every day with four. She sells the remainder at the farmers' market daily for \underline{\emph{\$2.5}} per fresh duck egg. \newline
       \textbf{Reversed Answer:} 27.5 \\
       \midrule
       \centering Fine-grained & \centering Sub-ability Question Generation & \centering\arraybackslash context, answer & 
        \textbf{New Question:} What are the detailed reasoning steps required to calculate how much in dollars Janet makes every day at the farmers' market? \newline
        \textbf{New Answer:} The solution involves 2 reasoning steps. [Step 1] calculates the number of eggs can be sold. [Step 2] calculate the money she earns.\\
     \bottomrule
    \end{tabular}
    }
    \caption{The reframing operations and examples for generating evolving instances.}
    \label{tab:operation_details}
    \vspace{-2mm}
\end{table*}
\siyuan{The overall illustration of our framework is presented in Figure~\ref{fig:system}. We first introduce different directions that we modify the contexts or questions of original instances along with their answers for newly evolving instances (see Section~\ref{sec2_taxonomy}). We employ a multi-agent system to facilitate collaboration on evolving instance generation and double-verification. (see Section~\ref{sec2_system})}
\begin{figure*}
    \centering
    \includegraphics[width=1.70\columnwidth]{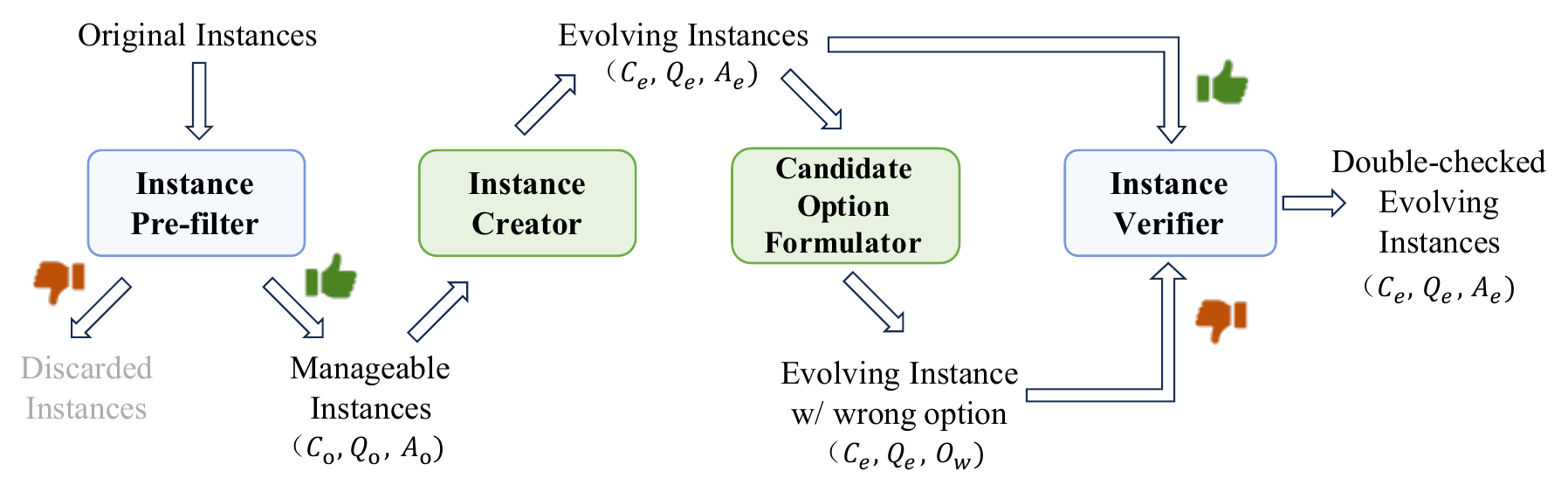}
    \caption{The workflow of our Multi-Agent Evolving Instance Generator system.}
    \label{fig:framework}
    \vspace{-1mm}
\end{figure*}

\subsection{Evolving Instance Taxonomy}
\label{sec2_taxonomy}
\siyuan{An instance can be formulated as a triplet consisting of a context ($C$), a question ($Q$) and an answer ($A$). For tasks involving only a question and an answer, the context is designated as null.
Given an original evaluation instance $(C_o, Q_o, A_o)$, we either perturb the context $C_o$ or alter the questions $Q_o$, simultaneously forming the corresponding answer. We thereby reframe an evolving instance as $(C_e, Q_o, A_e)$ or $(C_o, Q_e, A_e)$, for scalable, robust and fine-grained evaluation.} 

\vspace{2mm}
\noindent\textbf{Scalable Evaluation} \siyuan{For scalable evaluation of evolving LLMs, we create various questions with their corresponding new answers based on the original instance to examine whether LLMs can generalize to diverse and increasingly challenging queries. 
Our approach includes the creation of alternative questions (\textit{Question Alternating}) that examine different facets of the original context, as well as more complex questions requiring additional reasoning steps (\textit{Question Complicating}). To maintain the accuracy of evolving instance, we conduct question generation without changing original contexts.}

\vspace{2mm}
\noindent\textbf{Robust Evaluation} \siyuan{For more robust evaluation of LLMs, we introduce various perturbations to the contexts of original instances to generate evolving instances. Specifically, we apply three perturbation strategies on the contexts as follows. (1) \textit{Context Paraphrasing}: paraphrasing the original context to obtain diverse formulations; (2) \textit{Context Noising}: adding noise by introducing irrelevant or adversarial sentences into the original context; (3) \textit{Polarity Reversing}: reversing the polarity or altering key details of the original context. The first two perturbations require maintaining the original answer labels while the third approach necessitates a corresponding answer change, offering a more rigorous test of the model's adaptability.} 

\vspace{1.5mm}
\noindent\textbf{Fine-grained Evaluation} We design fine-grained evaluation by generating \textit{sub-ability questions} to probe LLMs' problem-solving capabilities.
We focus on three explainability-related sub-abilities: (1) task planning capability that inquires about the details of planned reasoning steps, (2) implicit knowledge identification capability for recognizing underlying facts or rules, and (3) relevant context retrieval capability for extracting pertinent information from the given context to support its responses.

The detailed operations to reframe evolving instances and corresponding examples are in Table~\ref{tab:operation_details}.

\subsection{Multi-Agent Evolving Instance Generator}
\label{sec2_system}
\siyuan{To generate evolving instances and ensure their correctness, we design a multi-agent instance creator system, incorporating four key agents including an instance pre-filter, an instance creator, an instance verifier and a candidate option formulator. All agents are built upon GPT-4 to fulfill their roles. The system's workflow is presented in Figure~\ref{fig:framework}.}

\vspace{1.5mm}
\noindent\textbf{Instance Pre-Filter}
The instance pre-filter is designed to go through the original dataset to identify manageable instances that fall within GPT-4's capability to accurately answer. As the whole system is powered by GPT-4, this instance pre-filter can establish a correct foundation for subsequent operations and enhance the overall system's reliability. It takes the context and question of the original instance as inputs, prompting GPT-4 to predict the answer and compare its prediction with the reference answer. A two-shot chain of thought~\cite{wei2022chain} prompting setting is utilized to select manageable cases as $(C_o, Q_o, A_o)$.

\vspace{1.5mm}
\noindent\textbf{Instance Creator}
\siyuan{The instance creator agent is pivotal in generating different types of newly evolving instances $(C_e, Q_e, A_e)$. Given an original instance including a context, a question, an answer, and its task description like ``mathematical reasoning task'', the instance creator is prompted to either modify the contexts ($Q_e=Q_o$) or form new questions ($C_e=C_o$). We design different prompts for six pre-defined reframing operations as Table~\ref{tab:operation_details}. For operations without altering the answer ($A_e=A_o$), the instance creator is specifically prompted to maintain the original answer during the operation. For operations leading to new answers, the instance creator think step-by-step to infer its new answer after the reformulation of the context or question. This process adopts a one-shot prompting strategy for better understanding operation requirements. An example prompt for generating complex question is provided below. Prompts for other reframing operations are listed in Appendix~\ref{appenx_prompts}.}
\vspace{-1mm}
\begin{table}[!h]
\begin{tcolorbox}[colback=blue!5!white,colframe=black,width=0.49\textwidth,title={Prompt on Question Complicating}]
\small
You are an expert Question Creator. You will receive an instance of \{\underline{task description}\}, including a context, an original question and its answer. \vspace{1mm}\\
Your task is to generate a more complex question and its corresponding answer based on the given context, with the goal of incorporating additional reasoning steps beyond what is required by the original question and answer.
Please do not change the context but just edit the question and the answer. \vspace{1mm}\\
Please first generate the question. Then think step-by-step in one line to give an brief analysis of the question, Finally, directly present a short answer omitting the intermediate steps, in a single line.  \\

\textbf{Context}: \{\underline{context $C_o$}\} \\
\textbf{Original Question}: \{\underline{question $Q_o$}\} \\
\textbf{Original Answer}: \{\underline{answer $A_o$}\} \\
\textbf{Alternative Question}:
\end{tcolorbox}
\vspace{-4mm}
\end{table}
 
\vspace{1.5mm}
\noindent\textbf{Instance Verifier}
\siyuan{The primary function of the instance verifier agent is to validate the correctness of the newly evolving instance $(C_e, Q_e, A_e)$, i.e., whether its answer can correctly support the corresponding context and question. Since these evolving instances are auto-generated through a GPT-4 based agent, the inclusion of this verifier is essential to control our data quality. The instance verifier directly takes the context, question and answer of the new instance as inputs, and adopts a two-shot CoT prompting strategy. It utilizes both a correct and an incorrect demonstrations to avoid potential biases. The specific prompt used for instance verifier can be found in Appendix~\ref{appenx_prompts}.}

\vspace{1.5mm}
\noindent\textbf{Candidate Option Formulator}
\siyuan{The candidate option formulator aims to generate an incorrect answer option $O_w$ for each new context-question pair $(C_e, Q_e)$. It has two primary purposes: (1) mitigating the impact of GPT-4’s shortcut biases on data reliability. It enables the instance verifier to conduct a double-check which simultaneously confirms the validity of the previous-generated instance $(C_e, Q_e, A_e)$ and ascertains that the candidate option $O_w$ is unable to answer the context-question pair $(C_e, Q_e)$. (2) It additionally provides a standardized binary-choice assessment method for more accurate evaluation metrics. For new instances with fine-grained questions where their free-form answers are not easy to evaluate, we adopt this binary-choice evaluation. Specifically, it takes the context-question pair and the correct answer as inputs, and adopts a one-shot prompting strategy to output a wrong candidate option. The specific prompt is provided in Appendix~\ref{appenx_prompts}.} 


\begin{table*}[!ht]
    \centering
    \setlength\tabcolsep{3pt}
    \resizebox{0.9\textwidth}{!}{
    \begin{tabular}{c|c|cc|ccc|c@{\hspace{5pt}}c@{\hspace{5pt}}c|c}
    \toprule
    \multirow{2}{*}{Dataset} & \multirow{2}{*}{Manageable} & \multicolumn{2}{c|}{Scalable} & \multicolumn{3}{c|}{Robust} &  \multicolumn{3}{c|}{Fine-Grained} & \multirow{2}{*}{\ Total \ } \\
    & & Alternating & Complicating & Paraphrasing & Noising & Reversing &  Planning & Knowledge & Retrieval & \\
    \midrule
    GSM8K & 96/100 & 65 & 55 & 90 & 90 & 61 & 71 & / & / & 432\\
    CLUTRR & 96/100 & 88 & 78 & 76 & 80 & 72 & 69 & 81 & 64 & 608\\
    StrategyQA & 83/100 & / & 57 & / & / & / & 78 & 65 & / & 200\\
    BoolQ & 90/100 & 88 & 68 & 90 & 86 & 50 & / & / & 67 & 382\\
    \bottomrule
    \end{tabular}
    }
    \caption{Statistics of our evolving instances from four original datasets.}
    \label{tab:dataset_statistics}
    \vspace{-2mm}
\end{table*}
\vspace{1.5mm}
\noindent\textbf{System Workflow}
\siyuan{The systematic procedure of our system is delineated in the following steps. First, the instance pre-filter selects manageable instances $(C_o, Q_o, A_o)$ from the original evaluation set. From these, the instance creator creates new instances $(C_e, Q_e, A_e)$ across various evolving types, and the candidate option formulator subsequently generates an incorrect candidate option $O_w$ for each new context-question pair.
Then the instance verifier need to validate the correctness of both the new generated instance $(C_e, Q_e, A_e)$ and its corresponding incorrect alternative $(C_e, Q_e, O_w)$. Only those instances that pass this double-verification process, i.e., the generated instance is examined as correct while the alternative is incorrect, will be utilized for dynamic evaluation. For a detailed exposition of the underlying algorithms, refer to Appendix~\ref{algoright_design}}

\section{Experiments}
\subsection{Setup}
\noindent\textbf{Tasks and Datasets}
Using our benchmark self-evolving framework, we dynamically extend benchmark datasets of four different tasks, including mathematical reasoning, logical reasoning, commonsense reasoning and reading comprehension. Specifically, we utilize GSM8K~\cite{cobbe2021training}, CLUTRR~\cite{sinha2019clutrr}, StrategyQA~\cite{geva2021did}, and BoolQ~\cite{clark2019boolq} respectively for these four tasks, all involving multi-step complex reasoning. Detailed dataset descriptions are provided in Appendix~\ref{appen: dataset_description}.

We randomly select 100 instances from publicly available dev/test sets of each dataset\footnote{The instances from CLUTRR are selected within the clauses of length $k \leq 3$.}, and feed each instance into our multi-agent system to generate new evaluation instances of various reframing types. For GSM8K, CLUTRR and BoolQ, we generate new instances across all six types as in Table~\ref{tab:operation_details}. For StrategyQA without context, we only generate instances with complex and fine-grained questions. Specifically for fine-grained sub-abilities, we focus on the task planning ability for GSM8K and BoolQ, both task planning and implicit knowledge identification for StrategyQA, and all three sub-abilities for CLUTRR. Statistics of generated evolving datasets are summarized as Table~\ref{tab:dataset_statistics}.

\vspace{2mm}
\noindent\textbf{Examined LLMs}
We evaluate both closed-source models, ChatGPT and ChatGLM~\cite{zeng2023glm-130b}, and open-source models, LLama~\cite{touvron2023llama} and Mistral~\cite{jiang2023mistral}, using our evolving evaluation datasets. We compare their performance against on original datasets to demonstrate the effectiveness of our framework.
For closed-source models, we use gpt-3.5-turbo-1106 and chatglm-turbo versions, while for open-source models, we employ LLama2-70B-Chat and Mistral-7B-Instruct-v0.2. We also evaluate GPT-4 despite its involvement in generating evolving instances, to test whether they can also provide more scalable and robust evaluation for GPT-4.

\vspace{2mm}
\noindent\textbf{Task Formats}
We adopt two task formats tailored to different evaluation directions. For scalable and robust evaluations, we adhere to the original datasets' task types, and employ a two-shot CoT prompting strategy. For fine-grained evaluation, we create binary-choice questions featuring two options, A and B, among which one is randomly selected as the correct answer and the other, developed by our Candidate Option Formulator, is incorrect. We implement a zero-shot prompting approach for fine-grained evaluation.

\subsection{Overall Comparison}
\siyuan{
We first provide an overall assessment of LLMs with scalable and robust evaluations, leaving the fine-grained evaluation in Section~\ref{sec:sub_ability}.
We conduct scalable evaluation using instances with alternative and complex questions, while performing robust evaluation using instances with paraphrased, noised and reversed contexts, and compare their performance against on corresponding original instances. For fair comparison, the average performance of original instances involved in each evaluation type is reported as their original performance. 
Table~\ref{tab:evolving_evaluation} presents the main comparisons, with arrows indicate shifts from original to evolving evaluation results. We have following findings.}

\begin{table*}[!ht]
    \centering
    \resizebox{0.7\textwidth}{!}{
    \begin{tabular}{c@{\hspace{0.4cm}}|@{\hspace{0.4cm}}c@{\hspace{0.8cm}}c@{\hspace{0.8cm}}c}
    \toprule
    \multirow{2}{*}{Models} & Scalable Evaluation & Robust Evaluation  & Overall \\
    & \small (Evolving$\leftarrow$Original) & \small (Evolving$\leftarrow$Original) & \small (Evolving$\leftarrow$Original) \\
    \midrule
    \multicolumn{4}{c}{GSM8K} \\
    \midrule
    GPT-4 & 85.00 $\leftarrow$ 100.0 (-15.00) \ \ & 97.10 $\leftarrow$ 100.0 (- 2.90)   & 93.07 $\leftarrow$ 100.0 (\bf- 6.93) \\
    ChatGPT & 60.83 $\leftarrow$ 93.33 (-32.50) \ \ & 79.25 $\leftarrow$ 91.29 (-12.04)  & 73.13 $\leftarrow$ 91.97 (\bf-18.84) \\
    ChatGLM  & 42.50 $\leftarrow$ 66.67 (-24.17) \ \ & 62.66 $\leftarrow$ 67.22 (- 4.56) & 55.96 $\leftarrow$ 67.04 (\bf-11.08) \\
    LLama & 40.83 $\leftarrow$ 60.00 (-19.17) \ \ & 60.58 $\leftarrow$ 58.51 (+ 2.07)  & 54.02 $\leftarrow$ 59.00 (\bf- 4.98) \\
    Mistral & 27.50 $\leftarrow$ 41.67 (-14.17) \ \  & 35.27 $\leftarrow$ 39.42 (- 4.15) & 32.69 $\leftarrow$ 40.17 (\bf- 7.48) \\
    \midrule
    \multicolumn{4}{c}{CLUTRR} \\
    \midrule
    GPT-4 & 77.11 $\leftarrow$ 100.0 (-22.89) \ \ & 93.42 $\leftarrow$ 100.0 (- 6.58)  & 86.55 $\leftarrow$ 100.0 (\bf-13.45) \\
    ChatGPT & 65.66 $\leftarrow$ 83.13 (-17.47) \ \ & 78.51 $\leftarrow$ 82.02 (- 3.51)  & 73.10 $\leftarrow$ 82.49 ( \bf-9.39) \\
    ChatGLM  & 55.42 $\leftarrow$ 73.49 (-18.07) \ \ & 67.11 $\leftarrow$ 74.56 (- 7.45) & 62.18 $\leftarrow$ 74.11 (\bf -11.93) \\
    LLama & 47.59 $\leftarrow$ 36.14 (+11.45) \ \ & 36.40 $\leftarrow$ 33.77 (+ 2.63)  & 41.12 $\leftarrow$ 34.77 (+ 5.35) \\
    Mistral & 45.78 $\leftarrow$ 55.42 (- 9.64) \ \ & 50.00 $\leftarrow$ 53.95 (- 3.95)  & 48.22 $\leftarrow$ 54.57 (\bf- 6.35) \\
    \midrule
    \multicolumn{4}{c}{StrategyQA} \\
    \midrule
    GPT-4 & 98.25 $\leftarrow$ 100.0 (- 1.75) \ \ & / & 98.25 $\leftarrow$ 100.0 (\bf- 1.75) \\ 
    ChatGPT & 64.91 $\leftarrow$ 91.23 (-26.32)  \ \ & / & 64.91 $\leftarrow$ 91.23 (\bf-26.32) \\
    ChatGLM  & 66.67 $\leftarrow$ 73.68 (- 7.01) \ \ & / & 66.67 $\leftarrow$ 73.68 (\bf- 7.01) \\
    LLama & 78.95 $\leftarrow$ 75.44 (+ 3.51) \ \ & / & 78.95 $\leftarrow$ 75.44 (+ 3.51) \\
    Mistral & 77.19 $\leftarrow$ 73.68 (+ 3.51) \ \ & / & 77.19 $\leftarrow$ 73.68 (+ 3.51) \\
    \midrule
    \multicolumn{4}{c}{BoolQ} \\
    \midrule
    GPT-4 & 99.36 $\leftarrow$ 100.0 (- 0.64) \ \ & 97.35 $\leftarrow$ 100.0 (- 2.65)   & 98.17$\leftarrow$ 100.0 (\bf- 1.83) \\
    ChatGPT & 92.31 $\leftarrow$ 91.03 (+ 1.28) \ \ & 91.15 $\leftarrow$ 90.27 (+ 0.88) & 91.62 $\leftarrow$ 90.58 (+ 1.04) \\
    ChatGLM  & 86.54 $\leftarrow$ 89.10 (- 2.56) \ \ & 90.71 $\leftarrow$ 88.05 (+ 2.66)  & 89.01 $\leftarrow$ 88.48 (+ 0.53) \\
    LLama & 84.62 $\leftarrow$ 92.31 (- 7.69) \ \ & 91.60 $\leftarrow$ 91.60 (- 0.00)  & 88.74 $\leftarrow$ 91.88 (\bf- 3.14) \\
    Mistral & 76.92 $\leftarrow$ 80.13 (- 3.21) \ \ & 83.19 $\leftarrow$ 79.20 (+ 3.99) & 80.63 $\leftarrow$ 79.58 (+ 1.05) \\
    \bottomrule
    \end{tabular}
    }
    \caption{Comparison of evolving and original evaluation. Left-side values of the arrow are evolving results, while the right-side denotes original performance on respective instances. Values in parentheses are performance changes.}
    \label{tab:evolving_evaluation}
    \vspace{-2mm}
\end{table*}
\begin{enumerate}[label=(\arabic*), itemsep=0.5pt, leftmargin=16pt, parsep=0.5pt, topsep=0.5pt]
\item Overall, most models exhibit reduced performance in our scalable and robust evaluation compared to their original results across the first three tasks. This offers a more accurate measure of LLMs' capabilities, highlighting that original results potential overestimate their proficiency. The decline of all models except LLama in robust evaluation on GSM8K and CLUTRR indicating their limited robustness.
\item GPT-4, though achieving 100\% accuracy on manageable original instances, still experiences performance drops on our evolving instances. This is because these evolving instances are generated by GPT-4 given the original ones with correct answers, aiding model's reasoning. Yet, directly evaluating GPT-4 with evolving instances would uncover their limitations.
\item Our scalable and robust evaluation effectively expand the performance gap between models. Initially, GPT-4 and ChatGPT exhibit less than a 10\% accuracy difference on GSM8K, while this gap increased to 20\% under our evolved evaluation. On the BoolQ dataset where all models consistently perform well, our scalable evaluation further highlights their disparities. 
\item Our framework also widens performance discrepancies of the same model across various tasks. For example, while ChatGPT consistently achieves 80$\sim$90\% accuracy on four tasks, its proficiency notably diverges following scalable evaluation, only maintaining stable performance on BoolQ. Similarly, GPT-4 shows performance disparities after our evaluation, maintaining effectiveness on most datasets while showing a decrease on CLUTRR.
\end{enumerate}

\subsection{Analysis of Varied Reframing Operations}
\label{sec:analysis_of_reframing_op}
\begin{figure}[!ht]
    \centering
    \includegraphics[width=0.99\columnwidth]{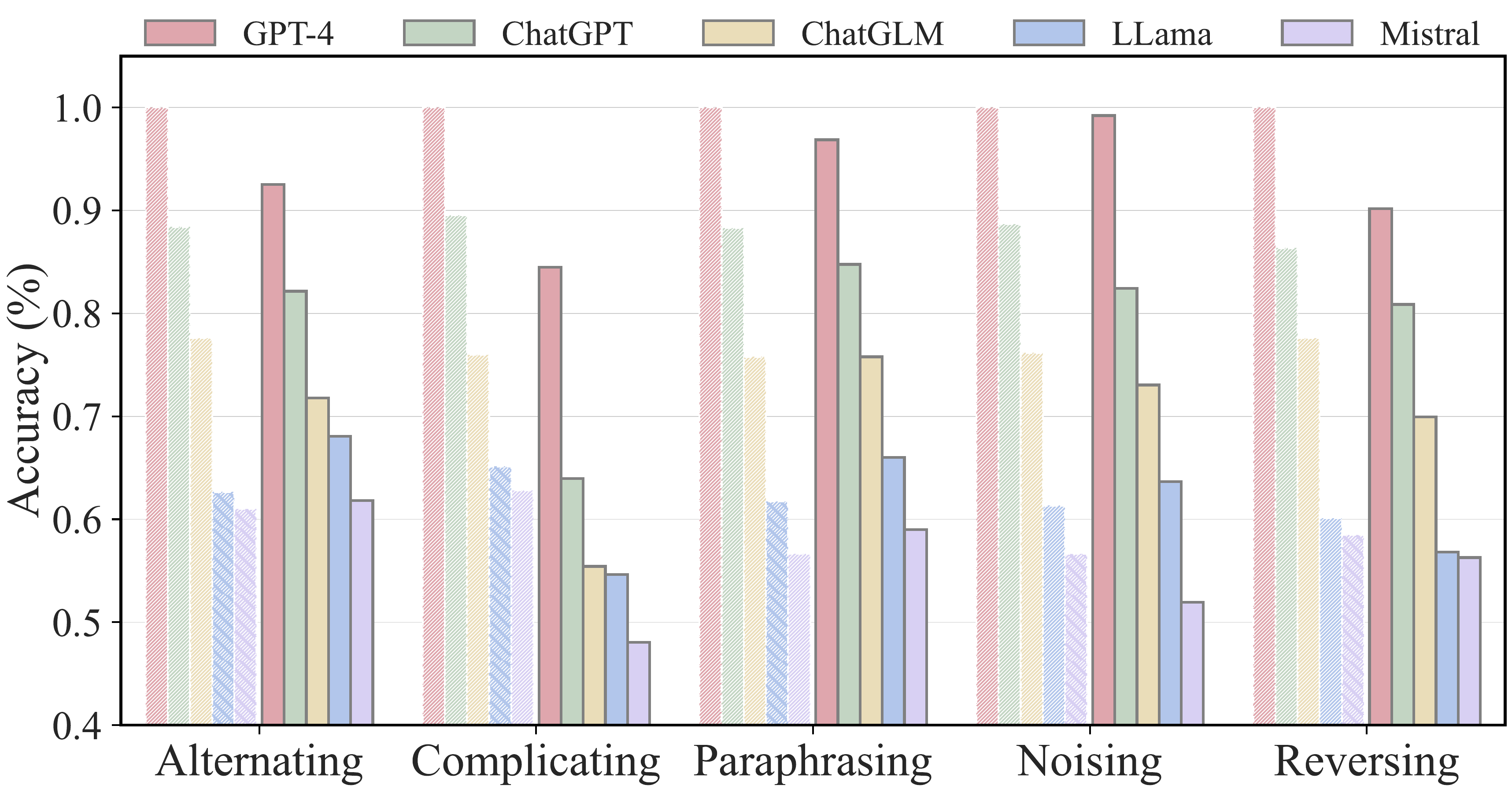}
    \caption{Comparison of evolving results using various reframing operations versus original results. Darker bars show accuracy for each operation across all datasets, with lighter bars ahead representing original accuracy.
    }
    \label{fig:results operations}
    \vspace{-1mm}
\end{figure}

To further assess the impact of various reframing operations on model evaluation, we gather results of each operation across all datasets and compare them with corresponding original results. Our analysis as detailed in Figure~\ref{fig:results operations} shows that among five reframing operations, \textit{question complicating} cause the most disruption to models, followed by \textit{polarity reversing} and \textit{question alternating}. In contrast,
\textit{context paraphrasing} and \textit{context noising} have a limited impact on model performance. These findings suggest that our framework primarily enhances the original benchmarks by highlighting the limitations of these LLMs regarding question generalizability and their susceptibility to adversarial attacks.

We further offer a perplexity-based analysis in Appendix~\ref{perplexity_analysis} demonstrating that our generated instances exhibit greater complexity and diversity than the original instances for dynamic evaluation, along with an error analysis in Appendix~\ref{error_analysis}.

\subsection{Further Analysis on Sub-Ability}
\label{sec:sub_ability}
\begin{figure}[t!]
    \centering
    \begin{subfigure}{0.83\columnwidth}
        \includegraphics[width=\columnwidth]{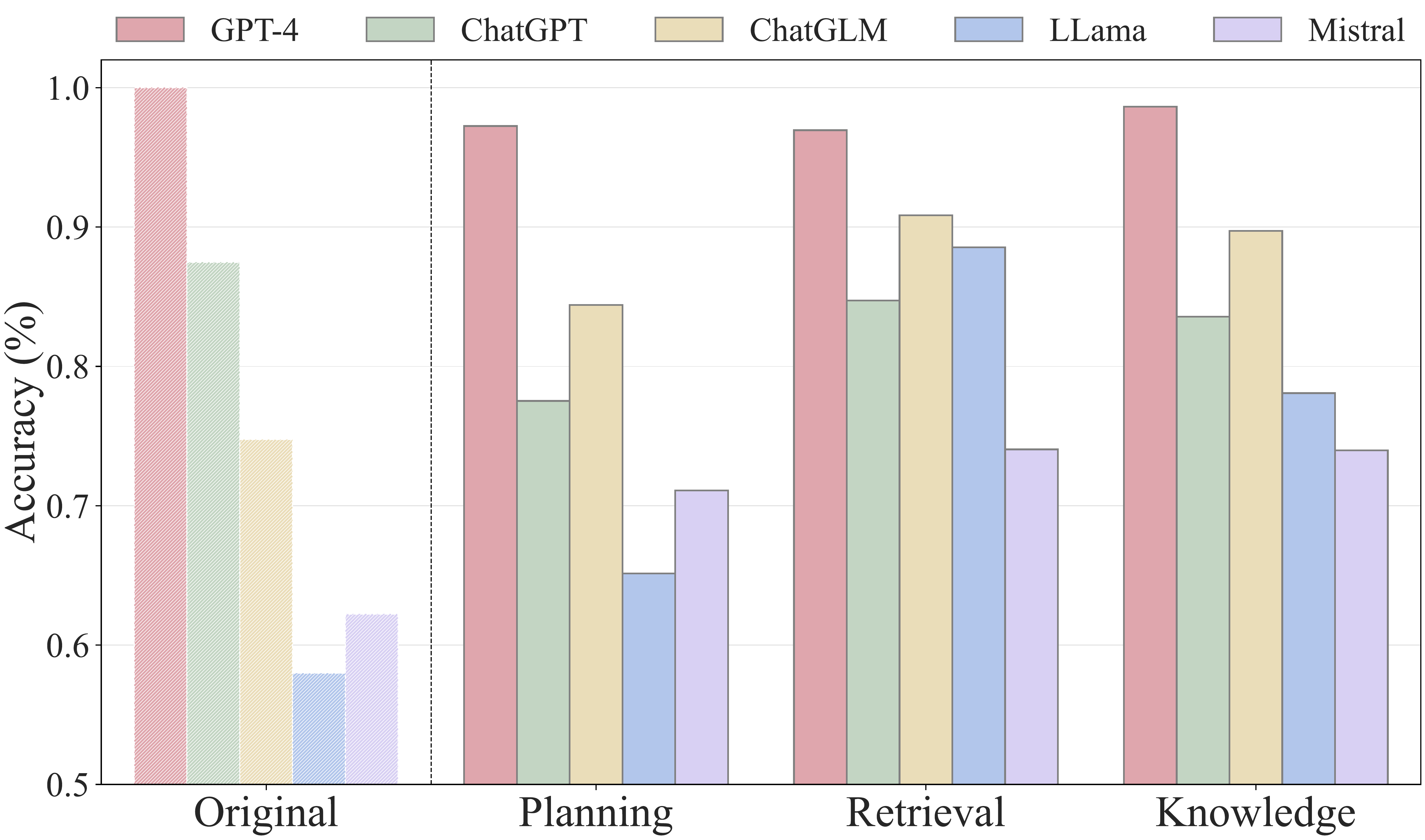}
        \caption{Biased results.}
        \label{fig:biased_result}
    \end{subfigure}
    \begin{subfigure}{0.82\columnwidth}
        \includegraphics[width=\columnwidth]{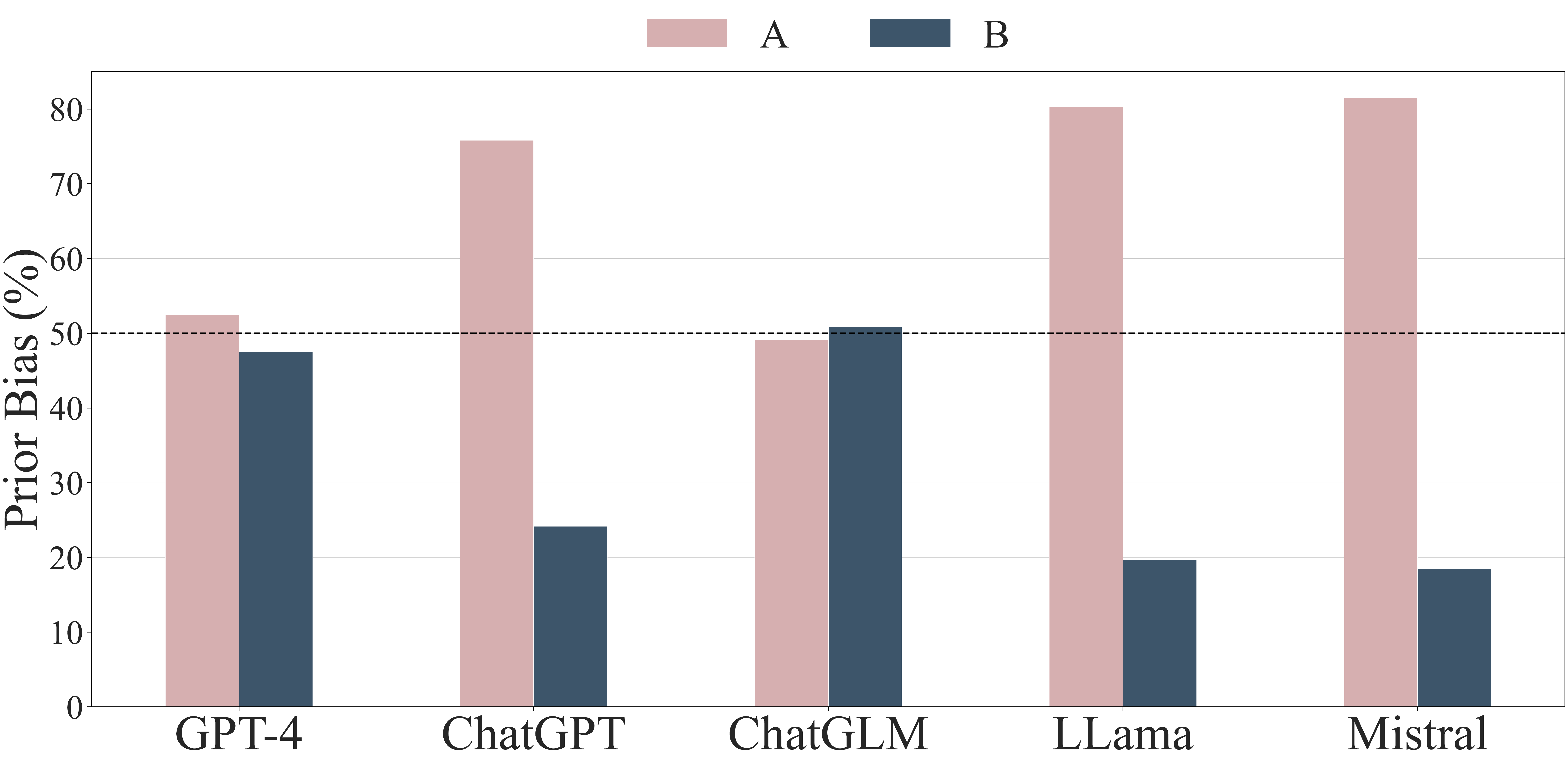}
        \caption{Selection Bias of Various LLMs.}
        \label{fig:bias}
    \end{subfigure}
    \begin{subfigure}{0.83\columnwidth}
        \includegraphics[width=\columnwidth]{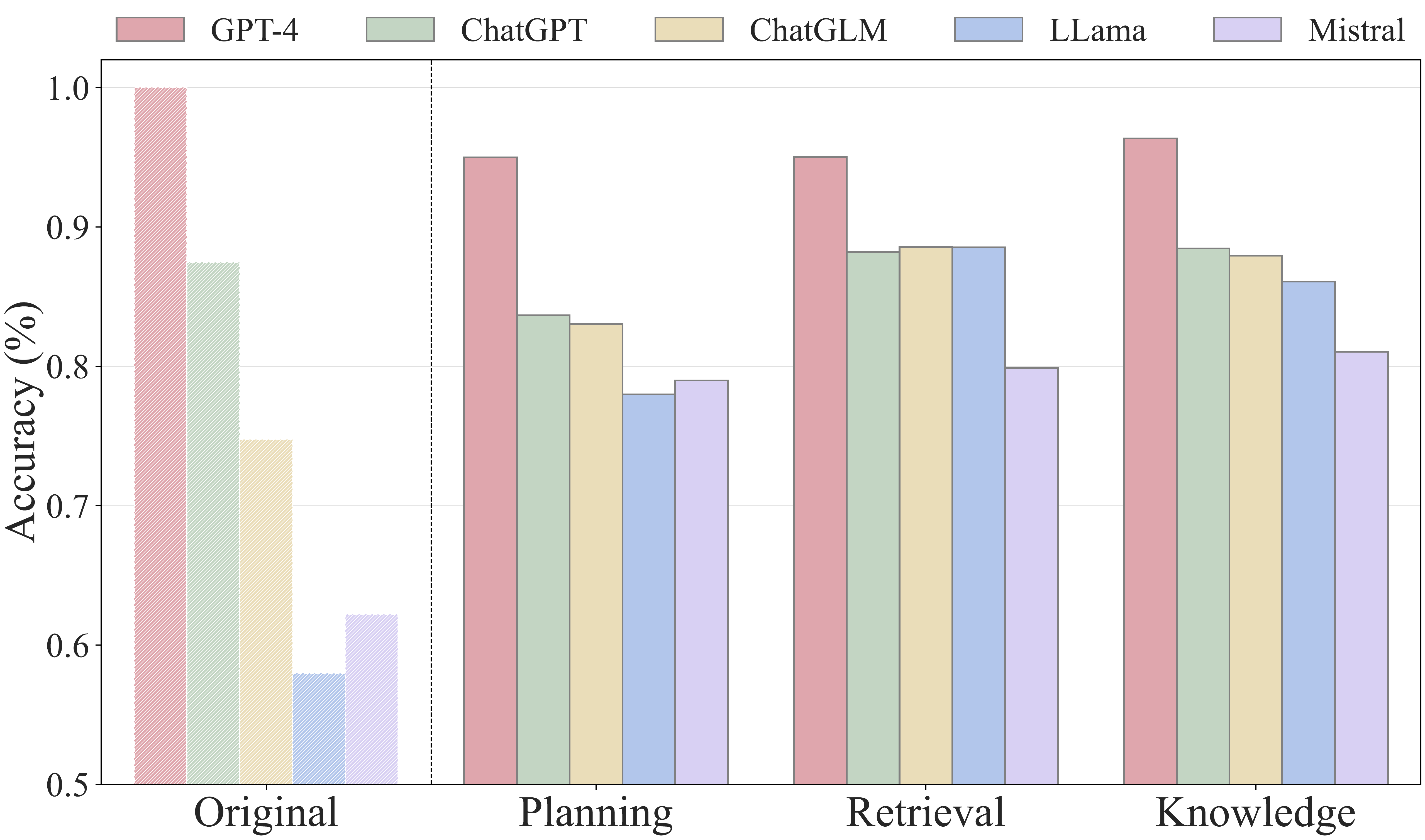}
        \caption{Debiased results.}
        \label{fig:debiased_result}
        \vspace{-1mm}
    \end{subfigure}
    \caption{Results of fine-grained sub-ability evaluation.}
    \label{fig:results subability}
    \vspace{-3mm}
\end{figure}

The fine-grained evaluation in our framework paves a way to dissect the sub-abilities of models. We aggregate the results of each sub-ability across all datasets and compare the models' rankings against their original ranking, as shown in Figure~\ref{fig:biased_result}.

\vspace{1.5mm}
\noindent\textbf{Substantial Discrepancy Between Original and Fine-grained Evaluations.} Our findings reveal that ChatGLM, initially lagging behind ChatGPT in original evaluation, surprisingly outperforms ChatGPT in all sub-ability evaluations. Upon scrutinizing ChatGPT's results, we observe a significant selection bias towards option 'A'. We hypothesize that such bias in certain LLMs impairs their decision-making, leading to poorer performance.

\vspace{1.5mm}
\noindent\textbf{Presence of Selection Bias in Certain LLMs.} Following~\cite{zheng2023large}, 
We estimate the prior prediction distribution of different LLMs on options ID 'A' and 'B'. The result in Figure~\ref{fig:bias} shows that ChatGPT, LLama, and Mistral exhibit a significant preference towards 'A', in contrast to the neutral stance of GPT-4 and ChatGLM.  
For a fair model evaluation, we utilize a bias calibrating method to obtain debiased results as shown in Figure~\ref{fig:debiased_result}, with methodologies for bias calculation and mitigation detailed in Appendix~\ref{appenx_selection_bias}.

\renewcommand{\thesubfigure}{\alph{subfigure}}
\begin{figure*}
    \centering
    \includegraphics[width=1.85\columnwidth]{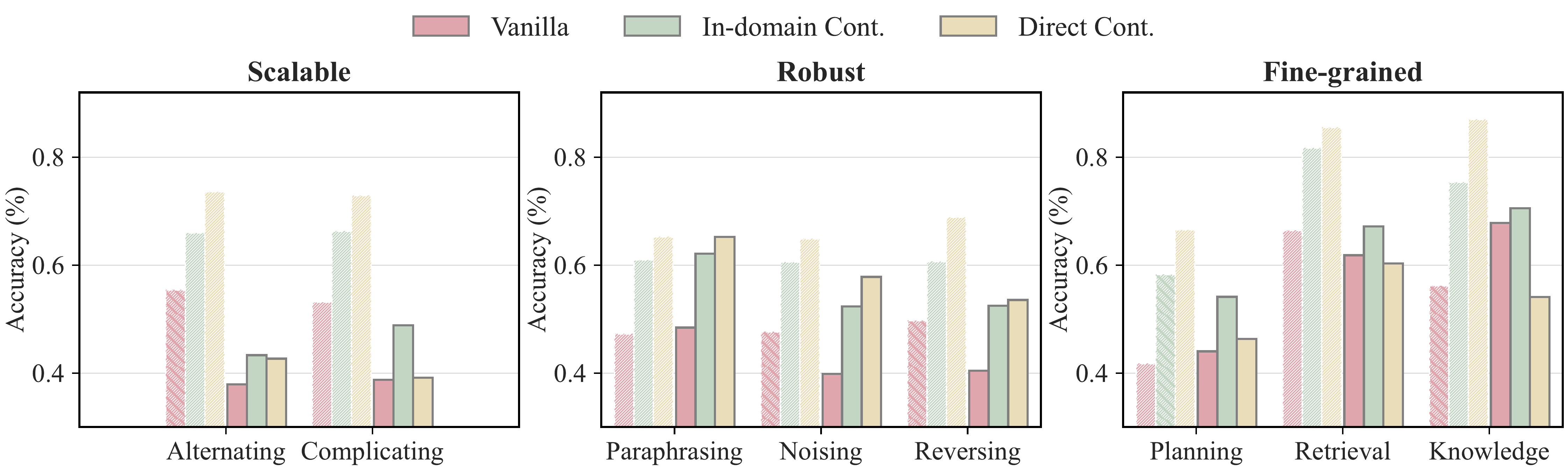}
    \caption{Comparison of LLama-2-7B-Chat models under different contamination conditions. ``Vanilla'', ``In-domain Cont.'' and ``Direct Cont.'' denotes the original model, the in-domain contaminated and direct contaminated models.}
    \label{fig:data contamination}
    \vspace{-2mm}
\end{figure*}

\vspace{1.5mm}
\noindent\textbf{Improvement Potential for Planning Ability} As Figure~\ref{fig:debiased_result}, GPT-4 consistently performs best across all three sub-abilities while Mistral showing the lowest performance. Among three sub-abilities, planning emerges as the weakest skill for all LLMs, highlighting a key area for further enhancements.


\subsection{Quality of Evolving Instances}
\noindent\textbf{Human Verification} To demonstrate the reliability of our dynamic evaluation, we sample a subset of our generated instances and conduct a human annotation~\footnote{Manually verified by the authors.} to assess their quality. Specifically, we randomly select five instances that are incorrectly answered by ChatGPT for each reframing operation across all datasets, with a total of 115 instances. Following human verification, 110 out of 115 instances (95.7\%) are deemed accurate, reinforcing the credibility of our evolving instances.

\vspace{2mm}
\noindent\textbf{Instance Filter Rate} Our system incorporates a pre-filter and a double-verification process to enhance the reliability of generated instances. The pre-filter discards nearly 9\% of the original instances that exceed the capabilities of GPT-4. Following this, the double-verification stage filters out approximately 24\% which initially processed correctly by GPT-4 (Detailed statistics are in Appendix~\ref{filtered dataset statistics}). This underscores the importance of our double-verification strategy on instance quality.

\subsection{Impact on Data Contamination}
To further analyze our framework's ability to mitigate data contamination, we design controlled experiments to simulate data contamination. We construct two instruction-tuning datasets: one simulating in-domain contamination by including portions of our evaluation benchmark's training set, and the other simulating direct contamination by incorporating both training and evaluation sets. We respectively use these two datasets to fine-tune LLama-2-7B-Chat (details in Appendix~\ref{appendix: data_contamination}).
We assess the original model and two fine-tuned models using original evaluation instances and our generated evolving instances, with results shown in Figure~\ref{fig:data contamination}.
\begin{itemize}[itemsep=0.5pt, leftmargin=16pt, parsep=0.5pt, topsep=0.5pt]
    \item Compared to the original model, both in-domain and direct contaminated models show notable improvement under original evaluation, revealing how data contamination can skew evaluations. In contrast, under our dynamic evaluation, the performance gap between contaminated models and the original model decreases, especially in scalable and fine-grained assessments. This indicates our framework's resistance against data contamination.
    \item In fine-grained evaluations, the in-domain contaminated model surpasses the original model, indicating that in-domain training can enhance task-related abilities. Yet, the direct contaminated model performs worse than the original, suggesting that memorizing original answers may hinder problem-solving in new contexts. This highlights the value of fine-grained evaluations in mitigating data contamination effects.
\end{itemize}


\section{Related Work}
\paragraph{LLMs Evaluation} \siyuan{Recent advanced large language models~\cite{openai2023gpt4,jiang2023mistral,touvron2023llama}, have shown remarkable performance across various tasks and sparked significant interest in their evaluation~\cite{chang2023survey, liu2023agentbench}. The evaluation for LLMs encompass automatic evaluation, human evaluation and LLM-based evaluation that respectively utilizes automatic metrics~\cite{liang2022holistic}, human preference~\cite{zheng2023judging} and LLM feedback~\cite{liu2023gpteval, zhang2023wider}. Automatic evaluation offers a most cost-effective approach for extensive and comprehensive assessments, necessitating diverse benchmark datasets, including task-specific benchmarks~\cite{yu2023metamath, holmes2023evaluating} and general benchmarks~\cite{hendrycks2020measuring, srivastava2023beyond}. 
However, with the rapid evolution of LLMs and potential data contamination issue, these static benchmarks tend to be inadequate for a thorough  assessment. To address this, our benchmark self-evolving framework dynamically updating instances via a multi-agent system, providing a dynamic solution for LLM evaluation.}

\vspace{1.5mm}
\noindent\textbf{Data Contamination} The expansion of training datasets for LLMs introduces a significant challenge of data contamination. This results in an in-domain distribution overlap with existing datasets, or even publicly available development and test sets, risking biased evaluation~\cite{sainz2023nlp}. 
Such contamination undermines the fairness and accuracy of benchmarks~\cite{zhou2023don}, casting doubt on whether high performance reflects true generalization or merely data memorization~\cite{biderman2023pythia}.
\citet{shi2023detecting} and \citet{golchin2023time} propose detecting contaminated data from benchmark for LLMs and use untainted data for evaluation. Besides, \citet{wei2023skywork} utilize perplexity as an evaluation metric on newly sampled data without additional annotations, yet this may not fully reflect models' capabilities. Our benchmark self-evolving framework can mitigate evaluation bias caused by data contamination.

\section{Conclusion}
In this study, we introduce a benchmark self-evolving framework that employs a multi-agent system to enhance existing benchmarks for a more scalable, robust and fine-grained evaluation on LLMs. 
Results indicate a general decline in LLM performance, alongside significant discrepancies in performance across various models and tasks. This highlights the effectiveness of our framework in providing a more accurate and comprehensive evaluation of LLMs, as well as distinguishing the capabilities of various models. Our study aids in selecting the most suitable LLMs for specific applications, and guide their further development.

\newpage
\section*{Limitations}
\paragraph{Limitation on benchmark coverage}
Due to computational limit, our dynamic evaluation study only explores four datasets across various textual tasks and select 100 instances from each dataset to construct nearly 1600 evolving instances. Our framework can flexibly generalize to other tasks and even different modalities for a broader analysis.

\paragraph{Limitation on examine LLMs}
We evaluate three closed-source (GPT-4, ChatGPT, ChatGLM) and two open-sourec LLMs (LLama, Mistral) using our crafted evolving instances to illustrate our scalable, robust and fine-grained evaluation. We acknowledge the limitation in the scope of LLMs, and will later provide further experiments on more LLMs.

\paragraph{Limitation on instance accuracy}
Despite incorporating a pre-filtering and double-verification procedure, our system, which is entirely powered by GPT-4, may inevitably generate a small number of instances with inaccuracies, as evidenced by human verification. This might result in less accurate assessments of LLMs.



\section*{Risks}
\paragraph{Introduction of Factual Errors}
For benchmark datasets containing factual information, such as BoolQ, our framework may generate counterfactual information to alter the key details of the original context during the polarity reversing operation. Such inaccuracies, if inadvertently used as learning material by the models, could negatively impact their performance and reliability.

\paragraph{Environmental Impact}
A significant risk associated with our methodology is the potential increase in environmental impact due to the extensive use of OpenAI's APIs for large language models. This is particularly concerning for benchmarks of substantial size, as the energy consumption and carbon footprint associated with generating evolving instances could be considerable.

\section*{Ethics Statement}
All data utilized in our benchmark self-evolving framework are sourced from publicly available datasets, including GSM8K, CLUTRR, StrategyQA and BoolQ. 
Our generated evolving instances for dynamic evaluation are also publicly released for usage and have been subjected to a thorough review by the authors. 
This setting guarantees transparency and reproducibility in our experiments, allowing other researchers to evaluate and expand upon our work.
Our benchmark-evolving framework is strictly limited to be used for instance generation that follow the ethical guidelines of the community. The authors emphatically denounce the use of our framework for generating inaccurate or harmful instances.

\bibliography{custom}

\appendix
\section{Details of Framework}
\subsection{Algorithm Design}
\label{algoright_design}
The pseudo-code for the algorithm of our multi-agent evolving instance setting system is presented in Algorithm~\ref{algorithm:1}. 

\begin{algorithm*}[ht]
\caption{Multi-Agent Evolving Instance Setter}
\begin{algorithmic}[1]
\REQUIRE An original evaluation instance $(C_o, Q_o, A_o)$, its task description $s$.
\ENSURE An evolving instance $(C_e, Q_e, A_e)$.
\STATE $(C_o, Q_o, A_o) \gets $ \texttt{Instance Pre-filter(}$(C_o, Q_o, A_o)$, $s$\texttt{)}
\STATE $(C_e, Q_e, A_e) \gets $ \texttt{Instance Generator(}$(C_o, Q_o, A_o)$, $s$\texttt{)}
\STATE \ $O_w \gets $ \texttt{Candidate Option Formulator(}$(C_e, Q_e, A_e)$, $s$\texttt{)}
\IF{\texttt{Instance Verifier}$(C_e, Q_e, A_e)$ \AND \NOT \texttt{Instance Verifier}$(C_e, Q_e, O_w)$}                
 \RETURN $(C_e, Q_e, A_e)$ 
\ELSE 
 \RETURN \emph{NULL}
\ENDIF    
\end{algorithmic}
\label{algorithm:1}
\end{algorithm*}

\subsection{Prompts of Multi-Agent Evolving Instance Setter}
\label{appenx_prompts}
Table~\ref{tab:prompt_alterate}, ~\ref{tab:prompt_phrase}, ~\ref{tab:prompt_noise}, ~\ref{tab:prompt_reverse}, ~\ref{tab:prompt_planning}, ~\ref{tab:prompt_retrieval} and~\ref{tab:prompt_knowledge} present prompts for Instance Generator of different reframing operations. Table~\ref{tab:prompt_verifier} presents the prompt for Instance Verifier. Table~\ref{tab:prompt_candidate} presents the prompt for Candidate Option Formulator.

\begin{table*}[!ht]
    \centering
    \vspace{2mm}
    \caption{Prompt on Question Alternating}
    \label{tab:prompt_alterate}
    \setlength\tabcolsep{3pt}
    \resizebox{1.0\textwidth}{!}{
    \begin{tcolorbox}[colback=blue!5!white,colframe=black,width=1.0\textwidth,title={Prompt for Instance Generator on Question Alternating}]
    \small
    You are an expert Question Creator. You will receive an instance of \{task description\}, including a context, a question and its answer. \\
    You are tasked with creating an alternative question to explore a different aspect of the original problem. \\
    Please do not change the context but just edit the question and the answer. \\
    Please first generate the question. Then think step-by-step in one line to give an brief analysis of the
    question, Finally, directly present a short answer omitting the intermediate steps, in a single line. \\
    
    Context: \{context $C_o$\} \\
    Original Question: \{question $Q_o$\} \\
    Original Answer: \{answer $A_o$\} \\
    Alternative Question: 
    \end{tcolorbox}
    }
\end{table*}

\begin{table*}[!ht]
    \centering
    \vspace{2mm}
    \caption{Prompt on Context Paraphrasing}
    \label{tab:prompt_phrase}
    \setlength\tabcolsep{3pt}
    \resizebox{1.0\textwidth}{!}{
        \begin{tcolorbox}[colback=blue!5!white,colframe=black,width=1.0\textwidth,title={Prompt for Instance Generator on Context Paraphrasing}]
        \small
        You are an expert Question Creator.You will receive an instance of \{task description\}, including a context, a question and its answer. \\
        Your task is to rephrase the given context in a short and easy-readable manner without summarizing or explaining. Confirm that the rephrased context do not change the answer to the original question. \\
        Simply output the rephrased context and do not output the original question. \\
        
        Context: \{context $C_o$\} \\
        Original Question: \{question $Q_o$\} \\
        Original Answer: \{answer $A_o$\} \\
        Alternative Context: 
        \end{tcolorbox}         
    }
\end{table*}

\begin{table*}[!ht]
    \centering
    \vspace{2mm}
    \caption{Prompt on Context Noising}
    \label{tab:prompt_noise}
    \setlength\tabcolsep{3pt}
    \resizebox{1.0\textwidth}{!}{
    \begin{tcolorbox}[colback=blue!5!white,colframe=black,width=1.0\textwidth,title={Prompt for Instance Generator on Context Noising}]
    \small
    You are an expert Question Creator. You will receive an instance of \{task description\}, including a context, a question and its answer. \\
    You are tasked with creating a new context by inserting irrelevant facts within the critical sentences of the original context. Make sure these facts shouldn't change the correct answer to the question. \\
    Simply output the rephrased context and do not output the original question. \\
    
    Context: \{context $C_o$\} \\
    Original Question: \{question $Q_o$\} \\
    Original Answer: \{answer $A_o$\} \\
    Alternative Context: 
    \end{tcolorbox}
    }
\end{table*}

\begin{table*}[!ht]
    \centering
    \vspace{2mm}
    \caption{Prompt on Polarity Reversing}
    \label{tab:prompt_reverse}
    \setlength\tabcolsep{3pt}
    \resizebox{1.0\textwidth}{!}{
    \begin{tcolorbox}[colback=blue!5!white,colframe=black,width=1.0\textwidth,title={Prompt for Instance Generator on Polarity Reversing}]
    \small
    You are an expert Question Creator. You will receive an instance of \{task description\}, including a context, a question and its answer. \\
    Your task is to generate a new context by altering key details in the original context. Ensure that the rest of the original context remains unchanged. The altered details should change the answer to the question. \\
    Please first output the rephrased context. Then give an one-line step-by-step analysis of the original question based on the new context. Finally, generate the corresponding direct answer in a newline. \\
    
    Context: \{context $C_o$\} \\
    Original Question: \{question $Q_o$\} \\
    Original Answer: \{answer $A_o$\} \\
    Alternative Context: 
    \end{tcolorbox}
    }
\end{table*}

\begin{table*}[!ht]
    \centering
    \vspace{2mm}
    \caption{Prompt on Planning}
    \label{tab:prompt_planning}
    \setlength\tabcolsep{3pt}
    \resizebox{1.0\textwidth}{!}{
    \begin{tcolorbox}[colback=blue!5!white,colframe=black,width=1.0\textwidth,title={Prompt for Instance Generator on Planning}]
    \small
    You are an expert Task Planner. You will receive an instance of \{task description\}, including a context, a question and its answer. \\
    Your task is to generate a new question and its corresponding answer, aiming to ask about the plan to solve the original question given the context. 
    Your new question can either inquire about all reasoning steps required or ask for the specific details about a certain (e.g., first, second, or last) step. \\
    Please first generate the question. Then think step-by-step in one line to give an brief analysis of the
    question, Finally, directly present a short answer omitting the intermediate steps, in a single line. \\
    
    Context: \{context $C_o$\} \\
    Original Question: \{question $Q_o$\} \\
    Original Answer: \{answer $A_o$\} \\
    Alternative Question: 
    \end{tcolorbox}
    }
\end{table*}

\begin{table*}[!ht]
    \centering
    \vspace{2mm}
    \caption{Prompt on Retrieval}
    \label{tab:prompt_retrieval}
    \setlength\tabcolsep{3pt}
    \resizebox{1.0\textwidth}{!}{
    \begin{tcolorbox}[colback=blue!5!white,colframe=black,width=1.0\textwidth,title={Prompt for Instance Generator on Retrieval}]
    \small
    You are an expert Relevant Context Retriever. You will receive an instance of \{task description\}, including a context, a question and its answer. \\
    Your task is to generate a new question and its corresponding answer, aiming to identify the relevant information from the given context necessary to solve the original question with the original answer. 
    Your answer must be exclusively from the given context, to contain all required information to solve the original question and cover the original answer. \\
    Please first generate the question. Then think step-by-step in one line to give an brief analysis of the
    question, Finally, directly present a short answer omitting the intermediate steps, in a single line. \\
    
    Context: \{context $C_o$\} \\
    Original Question: \{question $Q_o$\} \\
    Original Answer: \{answer $A_o$\} \\
    Alternative Question: 
    \end{tcolorbox}
    }
\end{table*}

\begin{table*}[!ht]
    \centering
    \vspace{2mm}
    \caption{Prompt on Knowledge}
    \label{tab:prompt_knowledge}
    \setlength\tabcolsep{3pt}
    \resizebox{1.0\textwidth}{!}{
    \begin{tcolorbox}[colback=blue!5!white,colframe=black,width=1.0\textwidth,title={Prompt for Instance Generator on Knowledge}]
    \small
    You are an expert Relevant Context Retriever. You will receive an instance of \{task description\}, including a context, a question and its answer. \\
    Your task is to generate a new question and its corresponding answer, aiming to ask about the implicit knowledge (e.g., facts, rules, commonsense, ...) required to solve the original question. 
    Your new answer should directly list all required implicit knowledge for the question. \\
    Please first generate the question. Then think step-by-step in one line to give an brief analysis of the
    question, Finally, directly present a short answer omitting the intermediate steps, in a single line. \\

    Context: \{context $C_o$\} \\
    Original Question: \{question $Q_o$\} \\
    Original Answer: \{answer $A_o$\} \\
    Alternative Question: 
    \end{tcolorbox}
    }
\end{table*}

\begin{table*}[!ht]
    \centering
    \vspace{2mm}
    \caption{Prompt for Instance Verifier}
    \label{tab:prompt_verifier}
    \setlength\tabcolsep{3pt}
    \resizebox{1.0\textwidth}{!}{
    \begin{tcolorbox}[colback=blue!5!white,colframe=black,width=1.0\textwidth,title={Prompt for Instance Verifier}]
    \small
    You are an expert Question-Answer Validator. You will receive an instance of \{task description\}, including a context, a question and its answer. \\
    Your task is to validate whether the answer is correct to solve the question given the context. \\
    Please think step-by-step in one line to analyze whether the answer is correct for the question and the context. Then give your final judgement with Yes or No in a newline. \\
    
    Context: \{context $C$\} \\
    Question: \{question $Q$\} \\
    Answer: \{answer $A$\} \\
    Judgement:
    \end{tcolorbox}
    }
\end{table*}

\begin{table*}[!ht]
    \centering
    \vspace{2mm}
    \caption{Prompt for Candidate Option Formulator}
    \label{tab:prompt_candidate}
    \setlength\tabcolsep{3pt}
    \resizebox{1.0\textwidth}{!}{
    \begin{tcolorbox}[colback=blue!5!white,colframe=black,width=1.0\textwidth,title={Prompt for Candidate Option Formulator}]
    \small
    You are an expert Candidate Option Generator. You will receive an instance of \{task description\}, including a context, a question and its answer. \\
    Your task is to modify the provided answer to generate a candidate option that wrongly answer the question given the context. \\
    
    Context: \{context $C$\} \\
    Question: \{question $Q$\} \\
    Answer: \{answer $A$\} \\
    Option:
    \end{tcolorbox}
    }
\end{table*}

\section{Experimental Analysis}
\subsection{Dataset Descriptions}
\label{appen: dataset_description}
\begin{itemize}[itemsep=0.2pt, leftmargin=9pt, parsep=0.5pt, topsep=0.5pt]
    \item GSM8K: a collection of grade school math problems in a free-form QA format, featuring diverse arithmetic and algebraic problems.
    \item CLUTRR: a synthesized free-form question answering dataset designed for evaluating logical reasoning over kinship relationships.
    \item StrategyQA: consists of crowdsourced yes/no questions that require implicit reasoning steps and commonsense strategies. The instances in StrategyQA consist solely of questions and answers, with their contexts being null.
    \item BoolQ: a reading comprehension dataset sourced from Google's Natural Questions, offers yes/no questions based on real Google searches paired with answers from Wikipedia articles.
\end{itemize}

\subsection{Instance Filtering Statistics}
\label{filtered dataset statistics}
Table~\ref{tab:filtered_dataset_statistics} shows the percentage of instances filtered by double-verification.
\begin{table*}[!ht]
    \centering
    \caption{Percentage (\%)  of instances filtered by double-verification.}
    \vspace{2mm}
    \label{tab:filtered_dataset_statistics}
    \setlength\tabcolsep{3pt}
    \resizebox{1.0\textwidth}{!}{
    \begin{tabular}{c|cc|ccc|c@{\hspace{5pt}}c@{\hspace{5pt}}c|c}
    \toprule
    \multirow{2}{*}{Dataset} &  \multicolumn{2}{c|}{Scalable} & \multicolumn{3}{c|}{Robust} &\multicolumn{3}{c|}{Fine-grained} & \multirow{2}{*}{\ Average \ }\\
    &  Alternating & Complicating & Paraphrasing & Noising & Reversing & Planning & Knowledge & Retrieval & \\
    \midrule
    GSM8K & 32.29\%& 42.71\%& 6.25\%& 6.25\%& 36.46\%& 26.04\%& / & / & 25.00\%\\
    CLUTRR & 8.33\%& 18.75\%& 20.83\%& 16.67\%& 25.00\%& 28.13\%& 15.63\%& 33.33\%& 20.83\%\\
    StrategyQA & / & 31.33\%& / & / & / & 6.02\%& 21.69\%& / & 19.68\%\\
    BoolQ & 2.22\%& 24.44\%& 0.00\%& 4.44\%& 44.44\%& / & / & 25.56\%& 29.26\%\\
    \midrule
    Average& 14.28\%& 29.17\%& 9.03\%& 9.12\%& 35.30\%& 20.06\%& 18.66\%& 29.45\%&23.69\%\\
    \bottomrule
    \end{tabular}
    }
\end{table*}

\subsection{Selection Bias Analysis}
\label{appenx_selection_bias}
In our discussion, \textit{selection bias} denotes the tendency that a model inherently assigns a higher probability to specific ID tokens, such as \textit{A} or \textit{B}, in multi-choice questions. The impact of \textit{selection bias} varies across different models and tasks, influencing the models' performance by reducing their robustness in handling multi-choice problems~\cite{zheng2023large}.
The permutation-based debiasing method, which averages the model's prediction distributions across various option permutations, theoretically eliminates selection bias. However, due to limited access to the prediction distributions of closed-source models, we employ a sampling approximation approach to estimate and mitigate selection bias.

To formalize our discussion, we use $C$ to denote the concatenation of context and question. Suppose an multi-choice problem consists of $n$ options, $id_i$ denotes the $i$th option ID in the default order, and $o_i$ denotes the $i$th option content in the default order, $i \in \{1, 2,..., n\}$. The option IDs remain in their default order in all multi-choice questions, while the option contents can be rearranged through different permutations to match with different option IDs. We introduce $I$ as a permutation of the set $\{1, 2, ..., n\}$, and $\mathcal{I}$ represents the set of all possible permutations $I$. We use $x^I$ to denote the concatenation of the default-ordered option IDs with the option contents permuted according to $I$.

We assume that given $C$ and $x^I$, the observed probability of the model selecting $id_i$ is $P_{\text{biased}}\left(id_i \mid C, x^I\right) $, which can be decomposed as:

\begin{align}
\label{eq:1}
Z_{x^I}^{-1} P_{\text{prior}}\left(id_i \mid C\right) P_{\text{debiased}}\left(o_{f_I(i)} \mid C, x^I\right)
\end{align}

where $Z_{x^I}^{-1}$ is the normalization factor, $P_{\text{prior}}\left(id_i \mid C\right)$ represents the prior probability of selecting $id_i$, $P_{\text{debiased}}\left(o_{f_I(i)} \mid C, x^I\right)$ represents the debiased prediction probability of $o_{f_I(i)}$, and $f_I(i)$ denotes the $i$th element in permutation $I$.

We assume $P_{\text{debiased}}$ is invariant to the ordering of options. By applying logarithms to both $P_{\text{biased}}$ and its decomposed expression, and then summing across all $I$ in $\mathcal{I}$ on both sides of the equation, we can simplify and compute $P_{\text {prior }}\left(id_i \mid C\right)$ as:

\begin{align}
    \operatorname{softmax}\left(\frac{1}{|\mathcal{I}|} \sum_{I \in \mathcal{I}} \log P_{\text {biased }}\left(id_i \mid C, x^I\right)\right)
\end{align}

Based on the estimated $P_{\text {prior }}\left(id_i \mid C\right)$ and Equation~\ref{eq:1}, we can compute the normalized $P_{\text {debiased }}\left(o_{f_I(i)} \mid C, x\right)$.

For our analysis, the $P_{\text {prior }}\left(id_i \mid C\right)$ and $P_{\text {debiased }}\left(o_{f_I(i)} \mid C, x\right)$ are computed independently for different datasets. Based on the correct option's ID, we categorize all multi-choice questions in a dataset into different permutation sets to support $x^I$. We calculate the frequency of the model selecting $id_i$ at different permutation sets to estimate $P_{\text {biased }}\left(id_i \mid C, x^I\right)$. 

\subsection{Dataset Perplexity Analysis}
\label{perplexity_analysis}
\begin{figure}
    \centering
    \includegraphics[width=0.98\columnwidth]{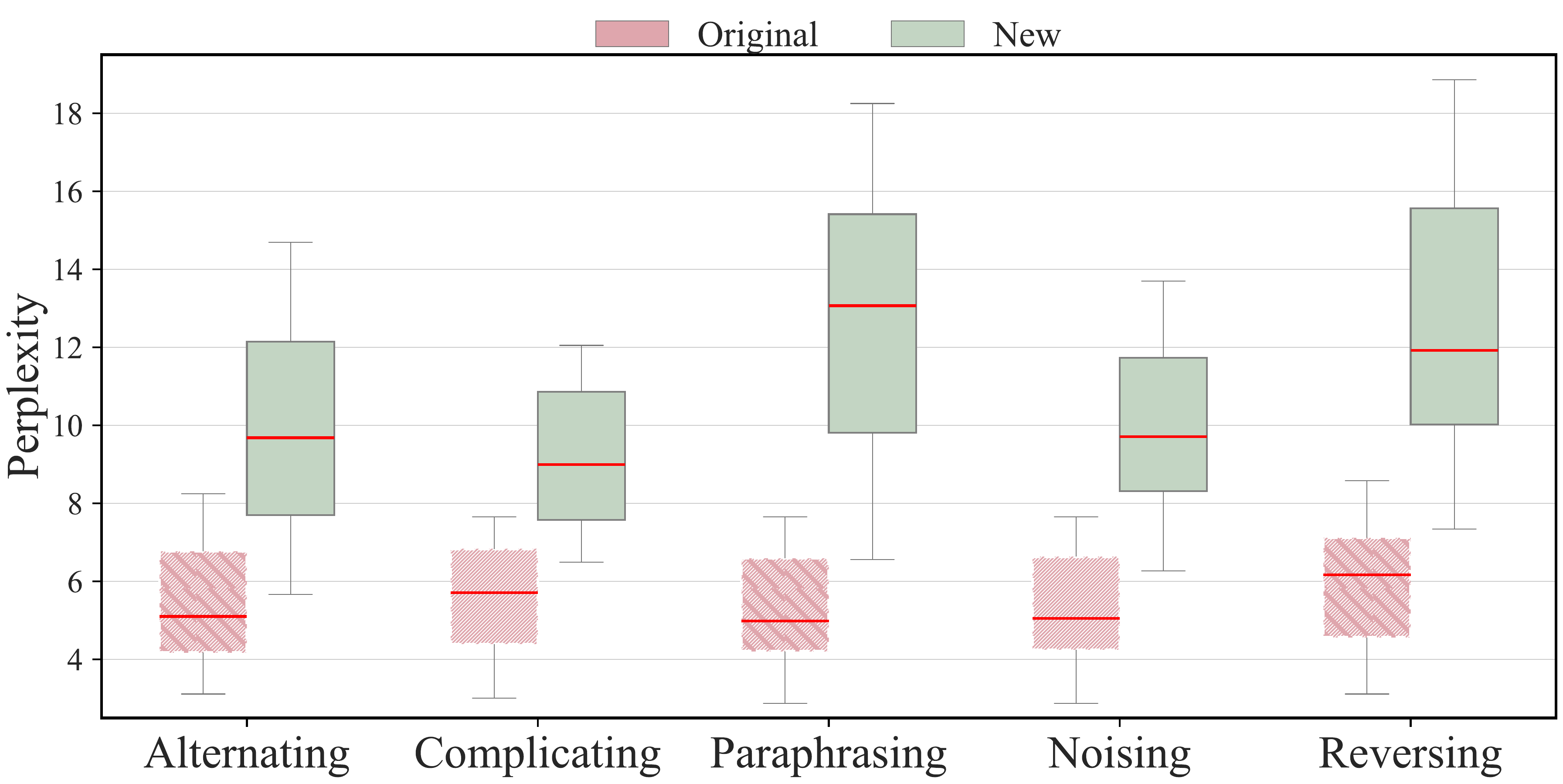}
    \caption{Perplexity comparison between original and reframed datasets.}
    \label{fig:perplexity}
\end{figure}

Perplexity is a metric that quantifies the complexity and the predictability of a dataset. By analyzing the perplexity of dataset, we can gain insights into the relative difficulty models may encounter during testing, as well as the diversity of information within the dataset.
In this analysis, we calculate the perplexity of newly evolving datasets derived by reframing the original GSM8K dataset and compared them with their original counterparts. The comparison results are presented in Figure~\ref{fig:perplexity}.

Our findings indicate that the datasets created through different reframing operations exhibit an increase in perplexity compared to the original instances. This indicates that the reframed instances are more complex and less predictable. These results, aligning with the experimental observations discussed in Section~\ref{sec:analysis_of_reframing_op}, suggest that our framework has the ability to generate instances with enhanced linguistic structure and diversity compared to the original instances.

\subsection{Error Analysis}
\label{error_analysis}
Tables~\ref{tab:dynamic_case},~\ref{tab:robust_case}and~\ref{tab:subability_case} present cases from GSM8K where ChatGPT successfully responds to the original instance but fails to answer the evolved version. 
The case in Table~\ref{tab:dynamic_case} introduces an additional reasoning step to the original instance, which causes ChatGPT to misinterpret the context and perform an incorrect calculation. In Table~\ref{tab:robust_case}, the evolved instance alters two data points from the original case, leading ChatGPT to neglect the critical information "4 times a week," resulting in an inaccurate response. The example in Table~\ref{tab:subability_case} demonstrates that, although ChatGPT correctly calculates the total number of hours the candle will burn in the original instance, it fails to address a direct question about the first reasoning step in the new instance.

By utilizing instances from our evolving framework, we can expose model's limitations more effectively, including its vulnerability to perturbations, diminished generalization capabilities on more complex problems, and inadequacies in addressing questions targeting specific sub-abilities.

\begin{table*}[!ht]
    \centering
    \caption{Instance Before and After Question Complicating}
    \vspace{2mm}
    \label{tab:dynamic_case}
    \setlength\tabcolsep{3pt}
    \begin{tabular}{
    >{\raggedright\arraybackslash}p{0.5\linewidth}
    >{\raggedright\arraybackslash}p{0.5\linewidth}}
        \toprule
        \textbf{Original Instance} & \textbf{New Instance} \\
        \midrule
        \textbf{Context:} A robe takes 2 bolts of blue fiber and half that much white fiber. & \textbf{Context:} A robe takes 2 bolts of blue fiber and half that much white fiber. \\
        \textbf{Question:} How many bolts in total does it take? &
        \textbf{Question:} If a merchant has 20 bolts of blue fiber and 10 bolts of white fiber, how many robes can he make using all the fibers without any leftovers? \\
        \textbf{Answer:} 3 & \textbf{Answer:} 10 \\
        \textbf{Response:} It takes 2 + 1 = 3 bolts in total. & \textbf{Response:} The merchant can \hlerror{make 20/2 = 10 robes using the blue fiber}. He can \hlerror{make 10/2 = 5 robes using the white fiber}. So, he can make a total of 5 robes using all the fibers without any leftovers. \\
        \textbf{Judgement:} \checkmarkgreen & \textbf{Judgement:} \xmarkred \\
        \bottomrule
    \end{tabular}
\end{table*}

\begin{table*}[!ht]
    \centering
    \caption{Instance Before and After Polarity Reversing}
    \vspace{2mm}
    \label{tab:robust_case}
    \setlength\tabcolsep{3pt}
    \begin{tabular}{
    >{\raggedright\arraybackslash}p{0.5\linewidth}
    >{\raggedright\arraybackslash}p{0.5\linewidth}}
        \toprule
        \textbf{Original Instance} & \textbf{New Instance} \\
        \midrule
        \textbf{Context:} Jim spends \hldel{2 hours} watching TV and then decides to go to bed and reads for half as long. He does this \hldel{3 times a week}. & \textbf{Context:} Jim spends \hladd{3 hours} watching TV and then decides to go to bed and reads for a third as long. He does this \hladd{4 times a week}. \\
        \textbf{Question:} How many hours does he spend on TV and reading in 4 weeks? &
        \textbf{Question:}  How many hours does he spend on TV and reading in 4 weeks? \\
        \textbf{Answer:} 36 & \textbf{Answer:} 64 \\
        \textbf{Response:} Jim spends 2 + (2/2) = 3 hours on TV and reading each time. So, in 4 weeks, he spends 3 * 3 * 4 = 36 hours on TV and reading. The answer is 36. & \textbf{Response:} Jim spends 3 + (3/3) = 4 hours on TV and reading each time. In 4 weeks, he spends \hlerror{4 * 4 = 16 hours} on TV and reading. The answer is 16. \\
        \textbf{Judgement:} \checkmarkgreen & \textbf{Judgement:} \xmarkred \\
        \bottomrule
    \end{tabular}
\end{table*}

\begin{table*}[!ht]
    \centering
    \caption{Instance Before and After Planning Operation}
    \vspace{2mm}
    \label{tab:subability_case}
    \setlength\tabcolsep{3pt}
    \begin{tabular}{
    >{\raggedright\arraybackslash}p{0.5\linewidth}
    >{\raggedright\arraybackslash}p{0.5\linewidth}}
        \toprule
        \textbf{Original Instance} & \textbf{New Instance} \\
        \midrule
        \textbf{Context:} A candle melts by 2 centimeters every hour that it burns. & \textbf{Context:} A candle melts by 2 centimeters every hour that it burns. \\
        \textbf{Question:} How many centimeters shorter will a candle be after burning from 1:00 PM to 5:00 PM? &
        \textbf{Question:}  What is the first reasoning step to calculate how many centimeters shorter a candle will be after burning from 1:00 PM to 5:00 PM?   \\
         & A. The first reasoning step is to measure the initial height of the candle. \\
         & B. The first reasoning step is to determine the total number of hours the candle will burn. \\
        \textbf{Answer:} 8 & \textbf{Answer:} B \\
        \textbf{Response:} The candle burns for 5-1=4 hours. So, it will be 2*4=8 centimeters shorter. The answer is 8. & \textbf{Response:} A \\
        \textbf{Judgement:} \checkmarkgreen & \textbf{Judgement:} \xmarkred \\
        \bottomrule
    \end{tabular}
\end{table*}

\subsection{Data Contamination Experiment Details}
\label{appendix: data_contamination}
We construct an instruction tuning dataset comprising 4,000 general instances from alpaca-gpt4-data~\cite{peng2023instruction} and additional 4,000 instances, with 1,000 each from the training sets of GSM8K, CLUTRR, StrategyQA and BoolQ. This dataset is used to fine-tune a model to simulate in-domain contamination. Furthermore, we incorperate 400 instances from the original benchmark into the instruction tuning dataset to fine-tune another model to simulate the direct contamination. 

\end{document}